# Termhood-based Comparability Metrics of Comparable Corpus in Special Domain

Sa Liu [1], and Chengzhi Zhang [1]

[1] Department of Information Management, Nanjing University of Science and Technology, Nanjing, China
`liusa321@163.com, zhangchz@istic.ac.cn`

**Abstract.** Cross-Language Information Retrieval (CLIR) and machine translation (MT) resources, such as dictionaries and parallel corpora, are scarce and hard to come by for special domains. Besides, these resources are just limited to a few languages, such as English, French, and Spanish and so on. So, obtaining comparable corpora automatically for such domains could be an answer to this problem effectively. Comparable corpora, that the subcorpora are not translations of each other, can be easily obtained from web. Therefore, building and using comparable corpora is often a more feasible option in multilingual information processing.

Comparability metrics is one of key issues in the field of building and using comparable corpus. Currently, there is no widely accepted definition or metrics method of corpus comparability. In fact, Different definitions or metrics methods of comparability might be given to suit various tasks about natural language processing. A new comparability, namely, termhood-based metrics, oriented to the task of bilingual terminology extraction, is proposed in this paper. In this method, words are ranked by termhood not frequency, and then the cosine similarities, calculated based on the ranking lists of word termhood, is used as comparability. Experiments results show that termhood-based metrics performs better than traditional frequency-based metrics.

## 1 Introduction

Parallel corpus which contains source documents and their translations plays an important role in multilingual information service [1], such as Cross-Language Information Retrieval (CLIR), and machine translation (MT). However, parallel corpus is scarce resources and not easy to be obtained in some under-resourced languages or special domains. Due to these shortcomings, building and using comparable corpora is often a more feasible option. It is obviously easier to find document collections with similar topics in multiple languages than to find parallel corpus [2]. The Web, with its vast volumes of data, offers a natural source for this. For example, bilingual website, and online Wikipedia, are very good resources for collecting and obtaining comparable data. Meanwhile, comparable data extracted from these resources can update with the increasing of the source data, and then more new words can be extracted easily and accurately. Therefore, building and using comparable corpora is becoming more and more important and urgent.

Comparability is the key concept in the research of comparable corpus. However, so far there has been no widely accepted definition of comparability. Different definitions or metrics methods of comparability might be given to suit various NLP tasks. In the task of machine translation, comparability is focused on distribution and quality of translated knowledge [3]. In multilingual terminology extraction, comparability is focused on distribution and quality of the vocabulary of translated forms [4].

So far, method of based word frequency list has been always used to measure corpus homogeneity and similarity between corpora [5]. This method is useful for measure corpus similarity in the respect of comparing the different language styles, however, this method perform badly in the task of bilingual term extraction. In our previous experiments, we verify this point.

A new comparability, namely, termhood-based metrics, especially used for comparability metrics of comparable corpus in special domain, is proposed in this paper. Experiments results show that this method performs better than traditional frequency-based metrics in the task of bilingual term extraction. The remainder of this paper is organized as follows. In section 2, related works are introduced. Then the termhood-based metrics is described in Section 3. Section 4 presents the experiment results and the proposed method is evaluated according to the task of bilingual terminology extraction in section 5. The paper is concluded with a summary and directions for future works.

## 2  Related Works

In this section, we review related works relevant to our research, including a brief review of building comparable corpus and comparability metrics of comparable corpus.

### 2.1  Building comparable corpus

Generally, comparable corpus is generated from news agencies or by crawling certain sites [2]. Talvensaari et al. built comparable corpora based on focused crawling [6]. Leturia et al. (2009) used search engine queries for collecting comparable corpora from the Internet [7]. Otero and L´opez exploited Wikipedia for collecting domain comparable corpora by using categories as topic restrictions [8].

In our previous experiments, we used three different Internet data source for collecting comparable corpus: querying bilingual domain keywords in search engine, exploiting the online encyclopedia-Wikipedia, and searching academic databases. At last, we choose data from academic databases as our experiment corpus for its suitable size and quality.

### 2.2 Comparability metrics of comparable corpus

In order to evaluate the quality of comparable corpus, we need some way to measure the degree of comparability of the comparable corpus. The most used metrics of comparable corpus are Chi-square statistics and word similarity. Leturia et al. used these two methods to measure the comparability of domain-specific comparable corpora obtained from the Internet by using search engine [2]. One method is calculating the cosine value between the vectors containing all the keywords of each corpus; the other is calculating Chi-square statistics for the most N frequent keywords (Top-N keywords). The ACCURAT (Analysis and evaluation of Comparable Corpora for Under Resourced Areas of Machine Translation) project used asymmetric Chi-square statistics to measure comparability [9]. The TTC (Terminology Extraction, Translation Tools and Comparable Corpora) project concentrate on two dimensions for comparability calculation: one is similarity between anchor texts in its own language; the other is dissimilarity between anchor points texts in two corpora [10].

The aforementioned works are based on word frequency lists. This kind of method is simple and effective for measure language style. However, this method performs poor between different domain corpora. In our previous investigation, we find this method can't distinguish different domain-specific corpus efficiently.

Li and Gaussier purposed a metrics of comparable corpus for bilingual lexicon extraction [11]. Given a comparable corpus P consisting of an English part $P_e$, and a French part $P_f$, the degree of comparability of P is defined as the expectation of finding the translation of any given source/target word in the target/source corpus vocabulary. This method needs a bilingual dictionary for mapping between two corpora. Thus the size and quality of dictionary may heavily affect the result of comparability measure.

In this paper, we propose termhood-based comparability metrics, according to bilingual terminology extraction task. It is noted that our method is oriented to bilingual terminology. It is in this point that our method is different to Li et al [11], which is oriented to bilingual lexicon. As for comparability of domain-specific comparable corpora, our method, based on termhood calculating, is more suitable to highlight terminology.

## 3 Termhood-based comparability metrics of comparable corpus

As for terminology extraction based on comparable corpus, comparability should concern the distribution and quality of terminology. Termhood is defined as degree of terminolgy to be term in a specific field. Quality of term can be measured by termhood and distribution of words be measured by ranking list of word weighting. So, we proposed termhood-based comparability metrics. In this method, words are ranked by termhood not frequency, and then the cosine similarities are calculated based on the ranking lists of word termhood. The similarity obtained is used as comparability of comparable corpus.

### 3.1 The Basic Idea

For calculating termhood-based comparability, the whole process we used is described as follows.

(1) Chinese-English domain comparable corpus collecting: comparable corpus we exploit in the experiment from two online academic databases (Chinese corpus from CNKI [12], English corpus from EBSCO[13]);

(2) Preprocessing: Chinese corpus is preprocessing and word lists from keywords (come from the abovementioned academic databases) and full-text words (come from full-text of document) are both obtained;

(3) Translating and processing: English corpus is translated into Chinese through Google translate [14]. Then the translated corpus is segmentated by ICTClAS[15], and finally word lists from keywords and full-text words are acquired;

(4) Termhood measure: Termhood of words is computed by using the method of corpus comparison after acquiring word frequency;

(5) Similarity calculating: Ranking the word list again based on termhood and calculating cosine similarity between vectors.

### 3.2 Key technology in the proposed method

In this section, key technology used in our proposed method is described in detail, including termhood measure and comparability of termhood-based metrics.

(1) Termhood measure by corpus comparison

It is observed that a true term is more outstanding (or peculiar) to its own subject domain than to a general domain or another field. Kit and Liu proposed a measure for mono-word termhood in terminology of such peculiarity and quantify it in terms of a word's ranking difference in a domain and background corpus [16]. We use this method to measure the termhood of terminology. We use People's Daily corpus of 1998 from January to June as background corpus and Library and information (LIS) corpus as domain corpus. Given a domain corpus D (with a vocabulary VD) and a background corpus B (with a vocabulary VB), the termhood of a word w is defined as formula (1).

Where r (w) is the ranking number of word w in a corpus in question, |VD|,| VB | is the size of domain and background corpus respectively. A word rank is normalized by the vocabulary size of its corpus in order to make the word ranks in two corpora comparable.

$$\text{Termhood}(w) = \frac{r_D(w)}{|V_D|} - \frac{r_B(w)}{|V_B|} \qquad (1)$$

(2) Comparability of termhood-based metrics in comparable corpus

According to the termhood, word lists are ranked in descending order. Then we calculate similarity of the new word list by vector space model [17]. The similarity obtained is used to be comparability of comparable corpus.

## 4 Experiments and Result Analysis

In this section, we first introduce experiments data used for comparability metrics. Then two different way of data processing are described in detail. At last, the experiment results are given and analyzed.

### 4.1 Data

The maximum and minimum of comparability is comparability of parallel corpus and comparability of non-comparable corpus respectively. In the experiments, we select three kinds of comparable corpus with different comparability, i.e. parallel corpus, comparable corpus, and non-comparable corpus. It is assumed that comparability of comparable corpus lies between parallel and non-comparable corpus. In our experiments, the domain of parallel corpus is Library and Information Science (LIS), obtained from abstracts of records from CNKI database; comparable corpus is also LIS domain, collected from two aforementioned academic databases. We build non-comparable corpus through combining Chinese corpus in domain of law and English corpus in domain of LIS. Table 1 describes basic information of corpus.

Table 1. Description of Experiment Corpus

| Experiment Corpus | Domain | number of tokens | |
|---|---|---|---|
| | | Chinese | English |
| parallel corpus | LIS | 81981 | 60841 |
| comparable corpus | LIS | 82024 | 60928 |
| non-comparable | Law & LIS | 29350 | 60928 |

Note: LIS denotes Library and Information Science.

### 4.2 Data processing

In order to verify the effectiveness of the proposed method, we compute the comparability of three kinds of comparable corpus, i.e. parallel corpus, comparable corpus and non-comparable corpus respectively. In the experiments, the comparability of each kind of comparable corpus is computed based on termhood-based and traditional frequency-based metrics respectively.

(1) Frequency-based metrics

Word was ranked based on their frequency in the corpus, and then comparability was calculated by cosine value between two vectors represented by words and their frequency. In the experiment, word frequency is normalized because there is some difference between size of Chinese corpus and English corpus. Experiments were carried out for six different corpus sizes, i.e. Top100, Top200, Top500, Top1000, Top2000 and Top5000 respectively.

(2) Termhood-based metrics

The comparability metrics based on termhood computes word termhood by corpus comparison method after word frequency statistic. Then words vectors are generated based on their termhood. Finally comparability was calculated by cosine value between two vectors represented by words and their termhood. In this method we also compute comparability metrics in six different corpus sizes, i.e. Top100, Top200, Top500, Top1000, Top2000, and Top5000.

Besides, keywords and all words were both used in our experiments for comparison. It should be noted that keywords come from the abovementioned academic database and all of words come from full-text of document after segment.

### 4.3 Experiments and Results

Fig.1 and Fig.2 are results of comparability based on keywords according to the two measurement methods.

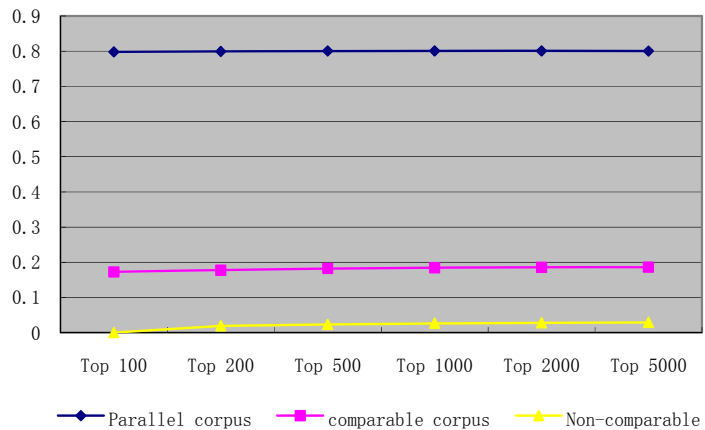

**Fig. 1.** Frequency-based metrics using keywords

Notice that both methods reflect the fact that comparability of parallel corpus > comparability of comparable corpus> comparability of non-comparable corpus. However, it is assumed that comparability of comparable corpus should be in the middle of the two; parallel, comparable, non-comparable, comparability of three kinds of corpus should present an even decreasing trends. In comparison, termhood-based approach reflects this point more obviously.

The performance of frequency-based method for all word is presented in Fig.3. As a whole, it reflects the fact that comparability of parallel corpus > comparability of non-comparable corpus> comparability of comparable corpus. This is against with our hypothesis. Moreover, this result is inconsistent with frequency-based method using keywords. Furthermore, we can also find that comparability of these kinds of corpus is very close to each other, all above 0.9. It is likely that there are so many noisy words that results are affected, further cause incorrect or incredible results.

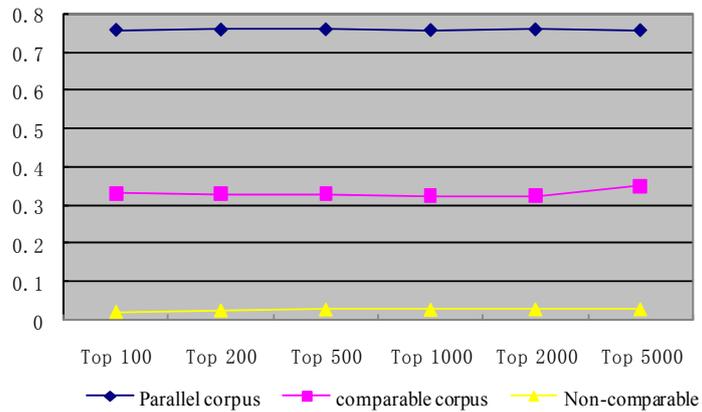

**Fig. 2.** Termhood-based metrics using keywords

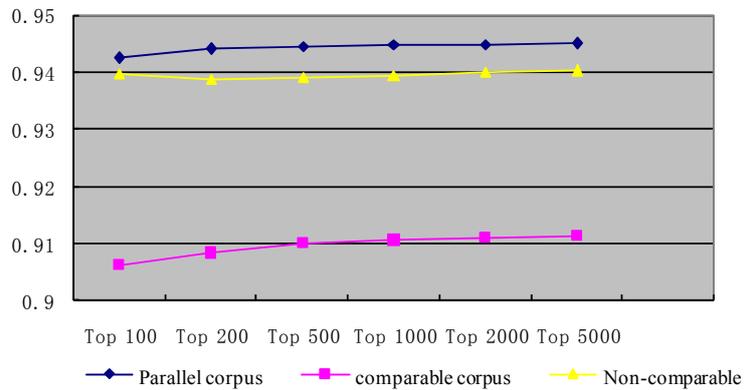

**Fig. 3.** Frequency-based metrics using all words

The performance of termhood-based method for all word is presented in Fig.4. Notice that comparability of parallel corpus > comparability of comparable corpus> comparability of non-comparable corpus; comparability of three kinds of corpus present an even decreasing trends. Furthermore, these results using all words are consistent with using only keywords.

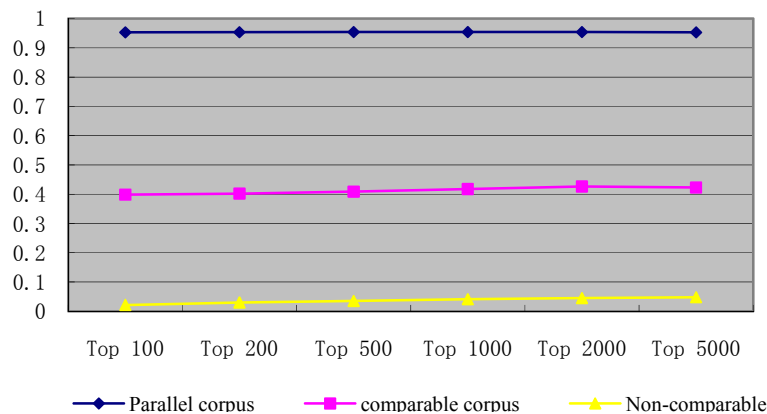

**Fig. 4.** Termhood-based metrics using all words

According to the above analysis, we can conclude that: according to reflecting the real comparable degree between corpora, termhood-based method is better than frequency-based method. Meanwhile, the results of termhood-based method using keywords are consistent with the situation of using full-text words. It also shows that termhood-based method is more stable and reliable than frequency-based method. Therefore, we can conclude that the performance of termhood-based metric method is better than frequency-based approach.

## 5 Evaluation

Furthermore, we verify the validity of the comparability measure of termhood-based metrics by the task of bilingual term extraction.

In this section, we need to learn whether comparable corpus with high comparability can generate bilingual terminology with high quality. Therefore, our experiment is designed to extract bilingual terminology from three corpora with different comparability. These corpora are parallel corpus, comparable corpus and non-comparable corpus respectively. Again, it is assumed that comparability of comparable corpus lies between parallel and non-comparable corpus. We expect that the higher corpus comparability is, the higher quality bilingual terminology we can obtain.

### 5.1 Methods of Bilingual Term Extraction and Evaluation

The method of terminology extraction in our experiment is one of the most popular methods, namely, context vector-based method [18], which includes the following steps.

1) Preprocessing. For Chinese corpus: segmentation and part of speech. For English: stemming and part of speech;
2) Generating candidate monolingual terminology;
3) Creating context vector of monolingual terminology based on co-occurrence statistics;
4) Translating Chinese context vector to Eng-lish through bilingual dictionary from LDC [19];
5) Computing similarity of context vector in singular space of English language.
6) Extracting terminology pairs of which similarity larger than the given threshold.
7) Evaluation of terminology quality.

We use Top@N method to evaluate the result of bilingual terminology extraction. That is, for every Top@N English terminology together with N candidate Chinese terms, if there is one is the right translation, we consider the result is correct. Here we take 10 for N. In the evaluation criteria of human judgments, if translation relation is completely correct, the score of matching will be 1; partly correct, score will be calculated by dice coefficient [20]; completely incorrect, score will be zero.

$$\text{Dice} = \frac{2 * overlaps}{\text{sum of number of tokens}} \quad (2)$$

### 5.2 Results and Analysis

We use two indicators for overall analysis, the overall similarity of terminology pairs and the overall degree of matching. Table 2 is the result of evaluation. Notice that the overall similarity of term pairs is increasing with the growth of comparability.

Table 2. Results of Evaluation

| Experiment Corpus | comparability | similarity | degree of matching | |
|---|---|---|---|---|
| | | | Machine | Human |
| non-comparable | 0.0478 | 0.2414 | 0.0480 | 0.0612 |
| comparable | 0.4226 | 0.3237 | 0.0395 | 0.0944 |
| parallel | 0.9527 | 0.3792 | 0.0566 | 0.0953 |

According to table 2, the overall degree of matching obtained by machine discrimination is not always increasing with the growth of comparability. In fact, machine discrimination is carried out with the help of dictionary of LDC. As far as we know, LDC's dictionary is a general dictionary which only includes about 80, 000 pairs bilingual lexicons, but the corpus in our experiments is corpus in special domain. So it is inevitable that there is some deviation because of limited size of bilingual dictionary. Therefore, human evaluation is very necessary. Notice that the overall degree of matching obtained by human judgments is increasing with the growth of comparability.

We can also find out that the overall degree of matching, no matter machine or human, is very low. This could well be due to the limited size and quality of bilingual dictionary, small size of our experimental corpus, or the bias caused by parameter settings in terminology extraction based on context vector method. At the same time, the overall similarity of terminology pairs is only related to terminology frequency. It is not affected by other factors. Therefore the result should be more reliable than the overall degree of matching.

In summary, we can conclude that the overall similarity of term pairs and the overall degree of matching is increasing with the growth of comparability. Accordingly, we can learn that high comparability of corpus generate high quality bilingual terminology. This also shows that our termhood-based method of comparability is effective in the application of bilingual terminology extraction.

## 6    Conclusions and Future Works

In this paper, we proposed termhood-based method to measure comparability of comparable corpus. Experiment results showed that this method performs better than traditional frequency-based method. It is likely to be that the candidate terms are ranked more reasonable because of constrain of termhood. It is possible that this method is less affected by common words, and it considers quality of term in special domain, so its performance is more stable and better in the task of terminology Extraction.

Experiments also show that results of comparability are more reliable when using keywords not full-text in the frequency-based method. This is because when using full-text data after preprocessing, there are so many common words that they influence the rank lists, further causing inaccurate results. Regardless of keywords or full-text data, the results are consistent in the termhood-based method. This again shows that termhood method has a better stability than frequency method.

Along the direction of our current work there are some directions for future works. One is to measure the effectiveness of termhood-base method in a more fine-grained comparable level. Another piece of work is to filter out the stop and common words after calculating word frequency, and then calculates the similarity of two words sequence; and finally we can make a comparison between this improved frequency-based method and termhood-based approach.

## Acknowledgement

This work is supported by National Natural Science Foundation of China under Grant No. 70903032.